\documentclass[letterpaper, 10 pt, conference]{ieeeconf}  

\IEEEoverridecommandlockouts                              

\usepackage{graphicx}
\usepackage{amsmath}
\usepackage{amssymb}
\usepackage{booktabs}
\usepackage[dvipsnames]{xcolor}
\usepackage{multirow}
\usepackage{hhline}
\usepackage[capitalize]{cleveref}
\usepackage{subcaption}
\title{\LARGE \bf
See Eye to Eye: A Lidar-Agnostic 3D Detection Framework for Unsupervised Multi-Target Domain Adaptation
}

\author{Darren Tsai, Julie Stephany Berrio, Mao Shan, Stewart Worrall, Eduardo Nebot
\thanks{Authors are with the Australian Centre for Field Robotics (ACFR) at the University of Sydney (NSW, Australia) E-mails: {\tt\small{\{d.tsai, j.berrio, m.shan, s.worrall, e.nebot}\}@acfr.usyd.edu.au}}%
}

\begin{document}

\maketitle
\thispagestyle{empty}
\pagestyle{empty}

\begin{abstract}

Sampling discrepancies between different manufacturers and models of lidar sensors result in inconsistent representations of objects. This leads to performance degradation when 3D detectors trained for one lidar are tested on other types of lidars. Remarkable progress in lidar manufacturing has brought about advances in mechanical, solid-state, and recently, adjustable scan pattern lidars. For the latter, existing works often require fine-tuning the model each time scan patterns are adjusted, which is infeasible. We explicitly deal with the sampling discrepancy by proposing a novel unsupervised multi-target domain adaptation framework, SEE, for transferring the performance of state-of-the-art 3D detectors across both fixed and flexible scan pattern lidars without requiring fine-tuning of models by end-users. Our approach interpolates the underlying geometry and normalises the scan pattern of objects from different lidars before passing them to the detection network. We demonstrate the effectiveness of SEE on public datasets, achieving state-of-the-art results, and additionally provide quantitative results on a novel high-resolution lidar to prove the industry applications of our framework. Our code and data are available at https://github.com/darrenjkt/SEE-MTDA.

\end{abstract}


\section{Introduction}

3D object detection is a fundamental task that allows perception systems to build an awareness of the environment with applications in mixed reality, collaborative robots, and autonomous driving. The release of large-scale, annotated public autonomous driving datasets \cite{sun2020scalability, geiger2012we, caesar2020nuscenes} has quickened the pace of 3D detection research, with novel architectures \cite{shi2020pv} consistently reaching greater heights on the leaderboards. Alongside software advances, lidar manufacturing has recently made outstanding progress with innovative sensor architectures and scan patterns; however, state-of-the-art (SoTA) detection architectures are not directly transferable to different lidar scan patterns \cite{yang2021st3d, yi2021complete}. 

\begin{figure}[t]
  \centering
   \includegraphics[width=0.87\linewidth]{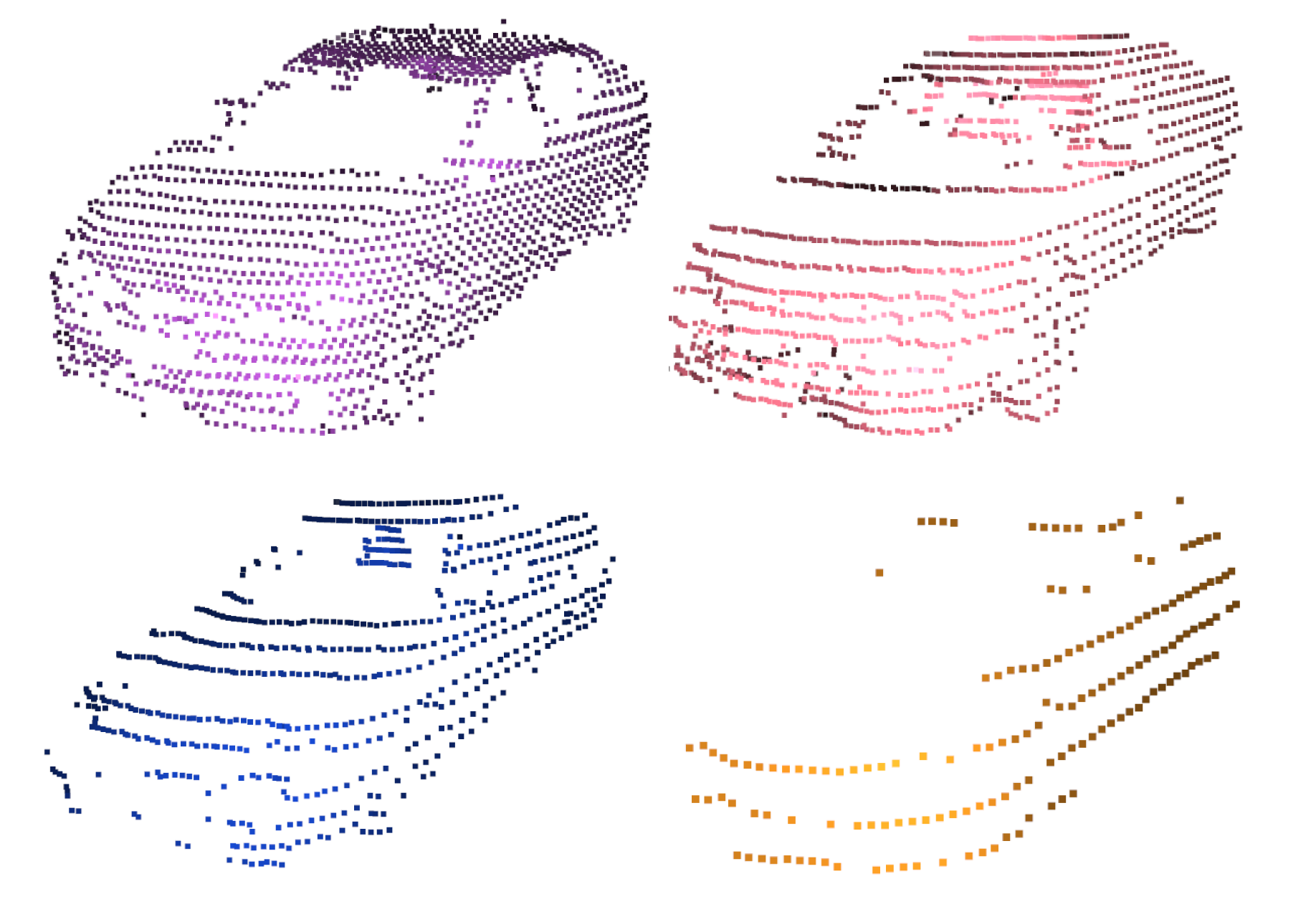}
   \caption{Sampling discrepancy of cars at 10m distance captured by different lidars. Top left: Baraja Spectrum-Scan™, an adjustable scan pattern lidar; Top right: Waymo Open Dataset \cite{sun2020scalability} (64-beam); Bottom left: KITTI Dataset \cite{geiger2012we} (64-beam); Bottom right: nuScenes Dataset \cite{caesar2020nuscenes} (32-beam). }
   \label{fig:cars}
   \vspace{-4mm}
\end{figure}

\begin{figure*}
  \centering
  \includegraphics[width=0.99\linewidth]{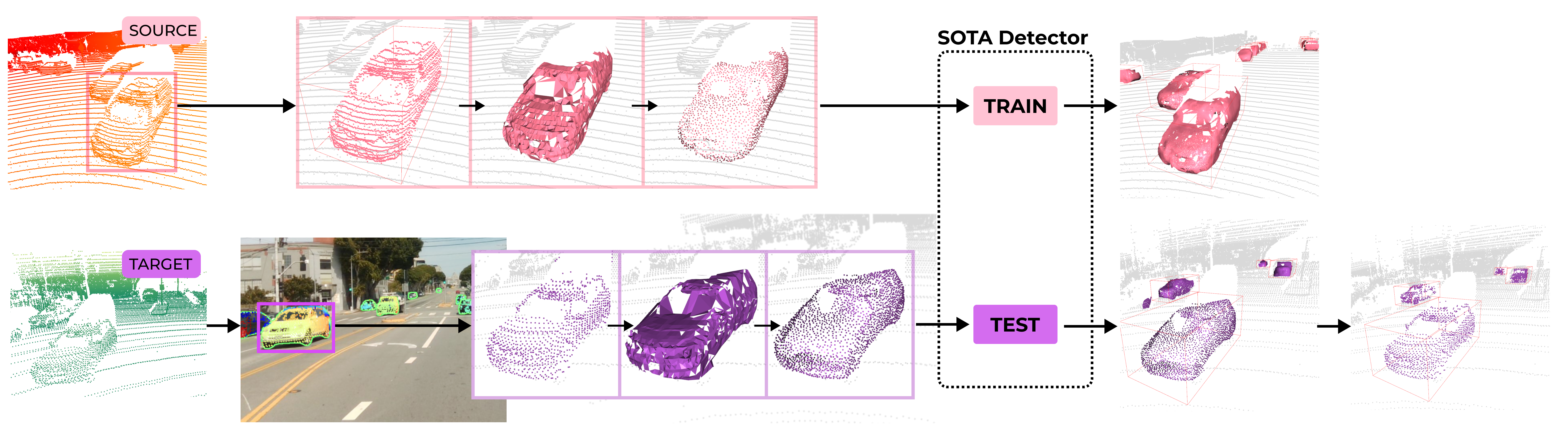}
  \caption{Overview of SEE, our lidar-agnostic unsupervised multi-target DA framework for 3D object detection. Our framework allows a 3D detector trained on one source domain to be used on multiple target domains without additional training. SEE is comprised of 3 phases: object isolation, surface completion (SC), and point sampling.}
  \label{fig:method_framework}
  \vspace{-4mm}
\end{figure*}

Each lidar has a distinct scan pattern, leading to a representation of objects specific to its scan pattern (\Cref{fig:cars}). This characteristic of lidars results in performance degradation when models trained for one lidar are tested on other types of lidars. In the supervised machine learning paradigm, this would require substantial effort to acquire manually annotated data for the new lidar scan pattern and to subsequently fine-tune the network. Approaches therefore that adapt a trained lidar domain to an unseen test domain are highly attractive for industry applications where it is costly to obtain labelled data.

Lidars often come in a variety of configurations, and some, like the Baraja Spectrum-Scan™, can adjust their scan pattern in real-time. Yang \textit{et al.} \cite{yang2021st3d} proposed a self-training approach using a memory-bank strategy to generate high-quality pseudo-labels. This approach is highly effective when transferring the performance of the SoTA from one labelled, fixed scan pattern lidar to a separate non-labelled, fixed scan pattern lidar. However, for lidars that have adjustable scan patterns, this would require fine-tuning a new model for each adjusted scan pattern, which is impractical. The motivation for this approach, therefore, is to explore a scan pattern agnostic representation of objects to enable a trained 3D detector to perform on any lidar scan pattern. 

We propose an unsupervised multi-target domain adaptation (MTDA) framework, SEE, that works on both fixed and adjustable scan pattern lidars without requiring fine-tuning a model for each new scan pattern. SEE enables the performance of SoTA models, trained on a single lidar dataset (source domain), to be adapted for multiple unlabelled lidar datasets of various, distinct scan patterns (target domains). Our approach focuses on transforming objects in their lidar-specific representation, to a scan pattern agnostic representation. We subsequently feed this into any SoTA detector and in so doing, we alleviate the bias towards a specific lidar's object representation.


We validate the efficacy of our framework with extensive experimentation using two SoTA detectors, SECOND \cite{yan2018second} and PV-RCNN \cite{shi2020pv}, on multiple public datasets and a manually labelled, novel lidar dataset. For instance, without domain adaptation, SECOND and PV-RCNN obtain an $\text{AP}_{\text{3D}}$ 0.7 IoU score of 11.92 and 41.54, respectively, when trained on the Waymo open dataset \cite{sun2020scalability} and tested on the KITTI dataset \cite{geiger2012we}. With SEE, the same models obtained a performance increase in 3D Average Precision (AP) of $11.92 \rightarrow 65.52$ and $41.54 \rightarrow 79.39$, which is an improvement over the state-of-the-art \cite{yang2021st3d} of 2.71 and 6.07 $\text{AP}_{\text{3D}}$, respectively. We additionally demonstrate the transferability of SECOND and PV-RCNN from Waymo and nuScenes to a novel high-resolution lidar, Baraja Spectrum-Scan™ to highlight SEE's ability to adapt to unseen scan patterns.

\noindent
{\bf Contributions.} Our main contribution, SEE, as shown in \Cref{fig:method_framework}, is a simple and effective framework that enables easy adoption of SoTA detectors for any kind of scan pattern. To the best of our knowledge, SEE is a pioneering approach for the MTDA task in the field of 3D detection. Through comprehensive experimentation, we demonstrate that SEE has the following advantages:
\begin{itemize}
\itemsep0em 
  \item {\bf Scan Pattern Agnostic}: Significant improvements achieved when transferring the performance of 3D detectors across lidars regardless if the lidar scan pattern is fixed or adjustable. 
  \item {\bf Unified Detector Comparison}: Any 3D detector can now be lidar-agnostic and easily benchmarked on multiple point cloud datasets.
  \item {\bf Detection Performance}: SEE achieves SoTA for the Waymo $\rightarrow$ KITTI task on all evaluated models.
  \item {\bf Lidar Innovation}: With SEE, novel scan patterns from innovative lidars can now be used with any detector; these previously needed expensive manual labelling, or significant hardware resources for fine-tuning.
\end{itemize}

\section{Related Works}
\noindent
{\bf 3D Object Detection.} The goal of 3D object detection is to obtain the class, location and orientation of all objects in the scene. Previous works can be categorised into lidar-based and multi-sensor methods. In lidar-based methods, point clouds are commonly converted to 2D representations such as birds-eye-view (BEV) \cite{yang2018pixor} or front-view (FV) \cite{chen2017multi}. These representations are then fed into 2D CNNs to regress bounding boxes and classes. Another popular lidar-based approach uses 3D CNNs and explores various representations of points such as voxels \cite{yan2018second}, spheres \cite{shi2020pv}, pillars \cite{lang2019pointpillars}, or learning point features directly \cite{pan20213d} using point operators \cite{qi2017pointnet++}. These features are transformed to a BEV feature map where 3D boxes are generated using sparse convolution or transformer architectures \cite{pan20213d}. 

For the multi-sensor approach, some methods \cite{chen2017multi} propose to fuse 2D point cloud representations (BEV and FV) with the raw image. Other works use image semantic segmentation \cite{vora2020pointpainting} or detection \cite{wang2019frustum} as priors for their proposed architectures. For experimentation, we validate our framework on 3D detectors, PV-RCNN \cite{shi2020pv} and a modified version of SECOND \cite{yan2018second}, named SECOND-IoU by \cite{yang2021st3d}.

\noindent
{\bf Unsupervised Domain Adaptation.} In supervised machine learning, the assumption is that the training and testing data are drawn from the same underlying distribution; however, this is often violated in practice. The goal of unsupervised MTDA is to address this distribution shift, to learn from a single labelled source domain in such a way that performs well on multiple unlabelled target domains. In this paper, ``source domain" refers to the labelled point cloud dataset used for training; ``target domain" refers to an unlabelled dataset with a different, unseen lidar. Domain-invariant feature representation is a popular line of work where divergence minimization is used to align the domains during training using adversarial training \cite{sankaranarayanan2018generate}, or divergence measures such as maximum mean discrepancy \cite{rozantsev2018beyond}, correlation alignment \cite{zhang2018unsupervised}, and contrastive domain discrepancy \cite{kang2019contrastive}. Other methods have opted to find a mapping from one domain to another to generate pseudo-target data for training with the known source labels \cite{hong2018conditional}.

Specific to DA for point clouds, recent works have addressed the domain shift for semantic segmentation \cite{wu2019squeezesegv2, yi2021complete, jaritz2020xmuda, luo2021unsupervised, langer2020domain}, weather interference \cite{xu2021spg}, time-of-day \cite{zhang2021srdan} and shape classification \cite{qin2019pointdan}. For the specific task of labelled source domain to unlabelled target domain for 3D object detection across distinct lidar scan patterns, research has been more sparse. Wang \textit{et al.} \cite{wang2020train} proposed a semi-supervised approach using object-size statistics of the target domain to resize training samples in the labelled source domain. A popular approach is the use of self-training \cite{you2021exploiting, yang2021st3d} with a focus on generating quality pseudo-labels using temporal information \cite{you2021exploiting} or an IoU scoring criterion for historical pseudo-labels \cite{yang2021st3d}. In particular, while ST3D \cite{yang2021st3d} has drastically improved the performance over previous works, it is a single target DA method, therefore requiring fine-tuning when adapting to new scan patterns. The Baraja Spectrum-Scan™ can adjust its vertical and horizontal angular resolution to achieve up to 1000 beams (far more than 64-beam Waymo or KITTI lidars). This flexibility enables users to define hundreds of scan-pattern configurations within a single lidar and adjust them in real-time. With such a lidar, ST3D would need to be fine-tuned for every adjustment of the scan pattern to obtain optimal performance. This thus limits the potential of having a flexible scan pattern in the first place. Conversely, we propose a novel MTDA framework that does not require fine-tuning for each new scan pattern and achieves SoTA performance without any labels or object statistics in the target domain.

\section{Method}
In this paper, we propose SEE. SEE is a framework for adapting a 3D detector trained on a labelled point cloud dataset with a fixed scan pattern to attain high performance on unlabelled point clouds with different scan patterns.

\subsection{Overview and notations}
Formally, we denote the point cloud as a set of unordered points $\mathbf{X}_i = \{p_j |^{N_j}_{j=1}\}$, where $p_j$ is the $j$-th point in the $i$-th point cloud $\mathbf{X}_i$ with $N_j$ number of points. Each labelled point cloud has a set of annotations $\mathbf{Y}_i = \{y_k|^{N_k}_{k=1}\}$ for $N_k$ objects, where $y_k$ is the $k$-th bounding box, parameterised by the center $(c_x, c_y, c_z)$, dimensions $(l, w, h)$ and rotation  $\theta_z$ around the world frame Z-axis. The point set of all object instances in the point cloud can be represented as $\mathbf{O}_i = \{o_k |^{N_k}_{k=1}\}$, where $o_k$ refers to the points of the $k$-th object. In this UDA scenario, we denote the source domain as $\chi^S = \{(\mathbf{X}^S_i, \mathbf{Y}^S_i)|^{N^S}_{i=1}\}$, where each $i$-th point cloud $\mathbf{X}^S_i$, has a corresponding annotation $\mathbf{Y}^S_i$ for all $N^S$ frames in the source domain. For the target domain, we are given a set of unlabelled point clouds $\chi^T = \{\mathbf{X}^T_i|^{N^T}_{i=1}\}$, where $\mathbf{X}^T_i$ is the $i$-th point cloud in the target domain with $N^T$ point clouds. 

SEE explicitly bridges the domain gap between distinct scan patterns by transforming only the object instances into a normalised representation. We call this resultant point cloud a semi-canonical domain where only the objects $\mathbf{O}_i^S$ and $\mathbf{O}_i^T$ reside inside a canonical domain. This can be represented by $\mathbf{M}^S_i = \psi(\mathbf{O}_i^S)$ and $\mathbf{M}^T_i = \psi(\mathbf{O}_i^T)$ where $\mathbf{M}^S_i$ and $\mathbf{M}^T_i$ represent the $i$-th semi-canonical point cloud. We train the 3D object detector $\rho(\mathbf{M}^S_i)$ on this semi-canonical source domain using labels $\mathbf{Y}^S_i$, and subsequently, in testing, apply it to the semi-canonical target domain $\rho(\mathbf{M}^T_i)$ to obtain bounding boxes which can then be projected back into the target domain $\mathbf{X}^T_i$. In this paper, the term ``rings" is used interchangeably with ``beams".

\subsection{Isolating the Object Point Cloud}
\label{sec:phase_isolating}
The first phase of SEE depicted aims to isolate the object points $\mathbf{O}_i^S$ and $\mathbf{O}_i^T$ in the point cloud. Where labels are available in the source domain, this can easily be accomplished for each $k$-th object by cropping the point cloud with their bounding box $y_k$. In the target domain, we utilize instance segmentation to isolate object instances $\mathbf{O}_i^T = \sigma (\mathbf{x}^T_i)$. 

\noindent
{\bf Instance segmentation.} Using image instance segmentation, we can obtain a mask that outlines the object shape within the detected bounding box. We first project the point cloud into the image plane using the homogenous transformation matrix $T_{\text{camera} \leftarrow \text{lidar}}$ and the camera matrix. Thereafter, in the image plane, we isolate the points within each mask. 

\begin{figure}[t]
  \centering
   \includegraphics[width=0.9\linewidth]{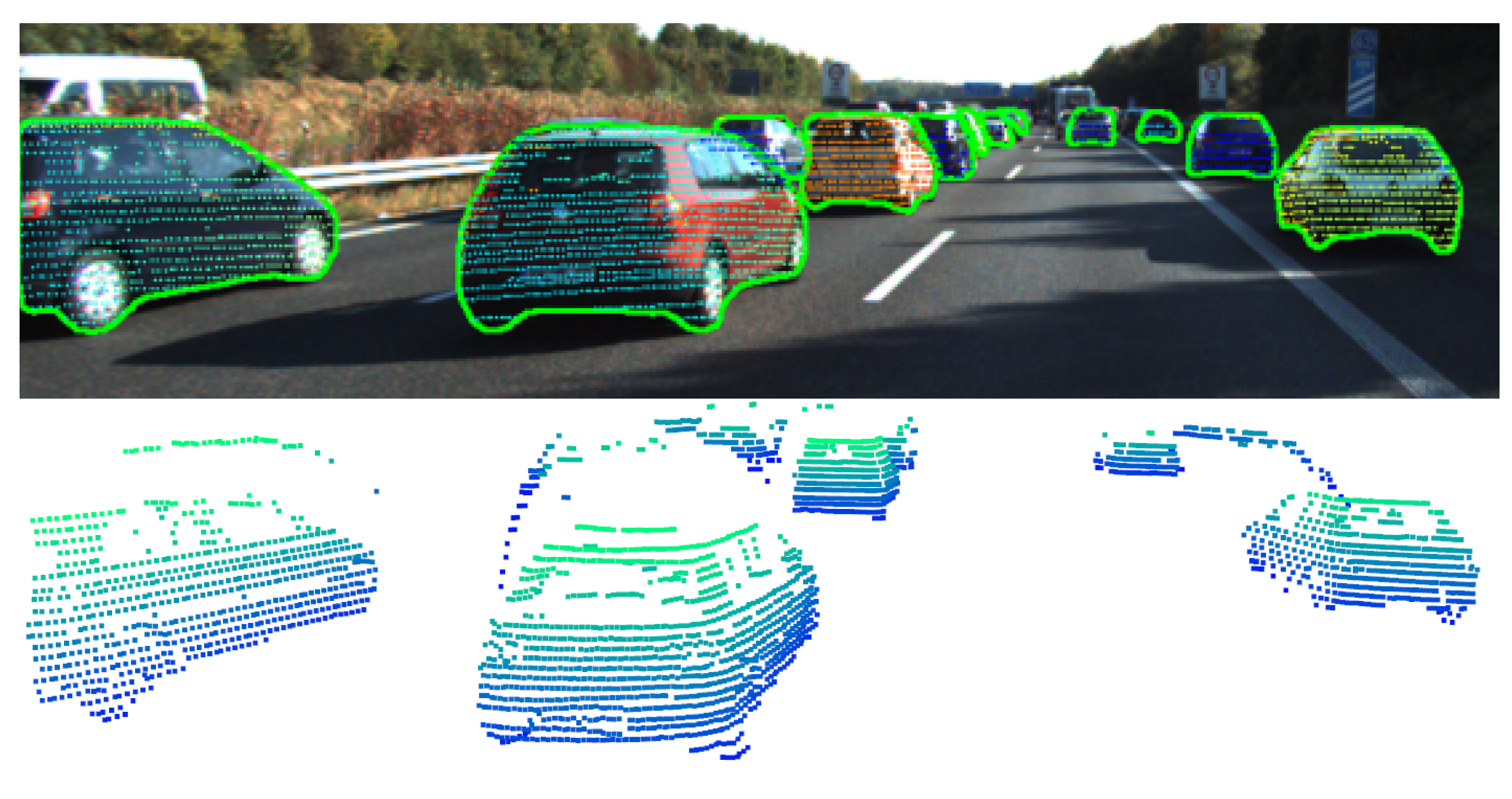}
   \caption{Background points due to viewpoint misalignment in KITTI dataset.}
   \label{fig:instance_seg}
   \vspace{-4mm}
\end{figure}

Instance segmentation masks allow for tight bounds on the object, giving better isolation of object points over bounding boxes; but this is not without its issues \cite{Berrio2021}. Viewpoint misalignment of camera and lidar, loosely fitted instance masks, as well as calibration errors \cite{tsai2021optimising} can lead to the inclusion of points within the mask that do not belong to the object, shown in \Cref{fig:instance_seg}. Additionally, due to the lack of depth in images, instance segmentation may occasionally pick up cars or people in posters and billboards \cite{gomez2020using}, which is not of interest to 3D object detection for autonomous driving. If these are transformed to the canonical domain and passed to the 3D detector for training, the model would learn the idiosyncrasies and failures specific to a particular instance segmentation model used for generating training data. To bypass this, we opt to use instance segmentation only in the target domain which in turn gives our framework flexibility in our choice of instance segmentation models. By using instance segmentation only in the target domain, we ensure that the source domain model is trained with clean data.

Our results show that for a well-calibrated lidar and camera pair with minimal viewpoint misalignment, a simple clustering strategy is sufficient to isolate object points $\mathbf{O}_i^T$ from the set of points in the mask $\mathbf{S}_i^T = \{s^T_k|^{N_k}_{k=1}\}$, where $s^T_k$ is the set of mask points for the $k$-th object.

\begin{figure}[t]
  \centering
  \begin{subfigure}{0.4\linewidth}
    \includegraphics[width=0.99\linewidth]{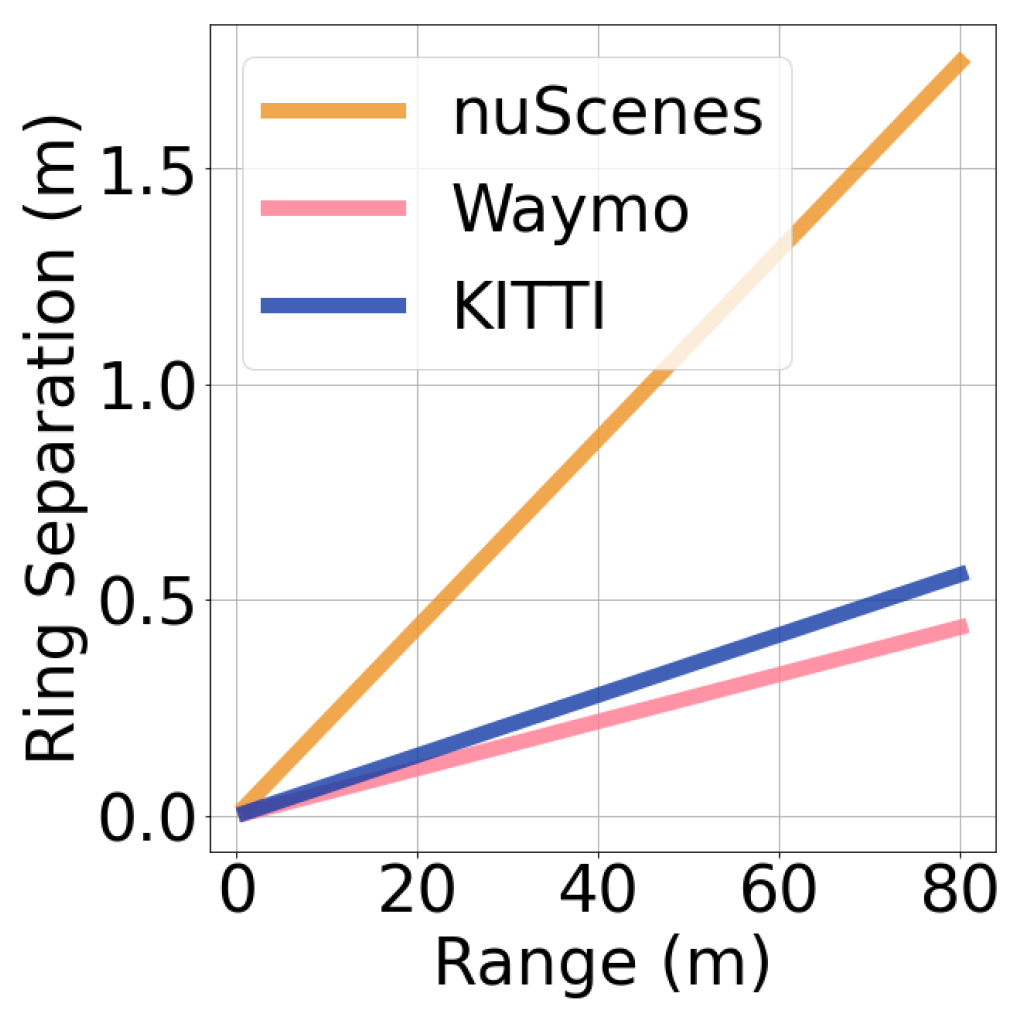}
    \caption{}
    \label{fig:ring_separation-a}
  \end{subfigure}
  \hfill
  \begin{subfigure}{0.57\linewidth}
    \includegraphics[width=0.99\linewidth]{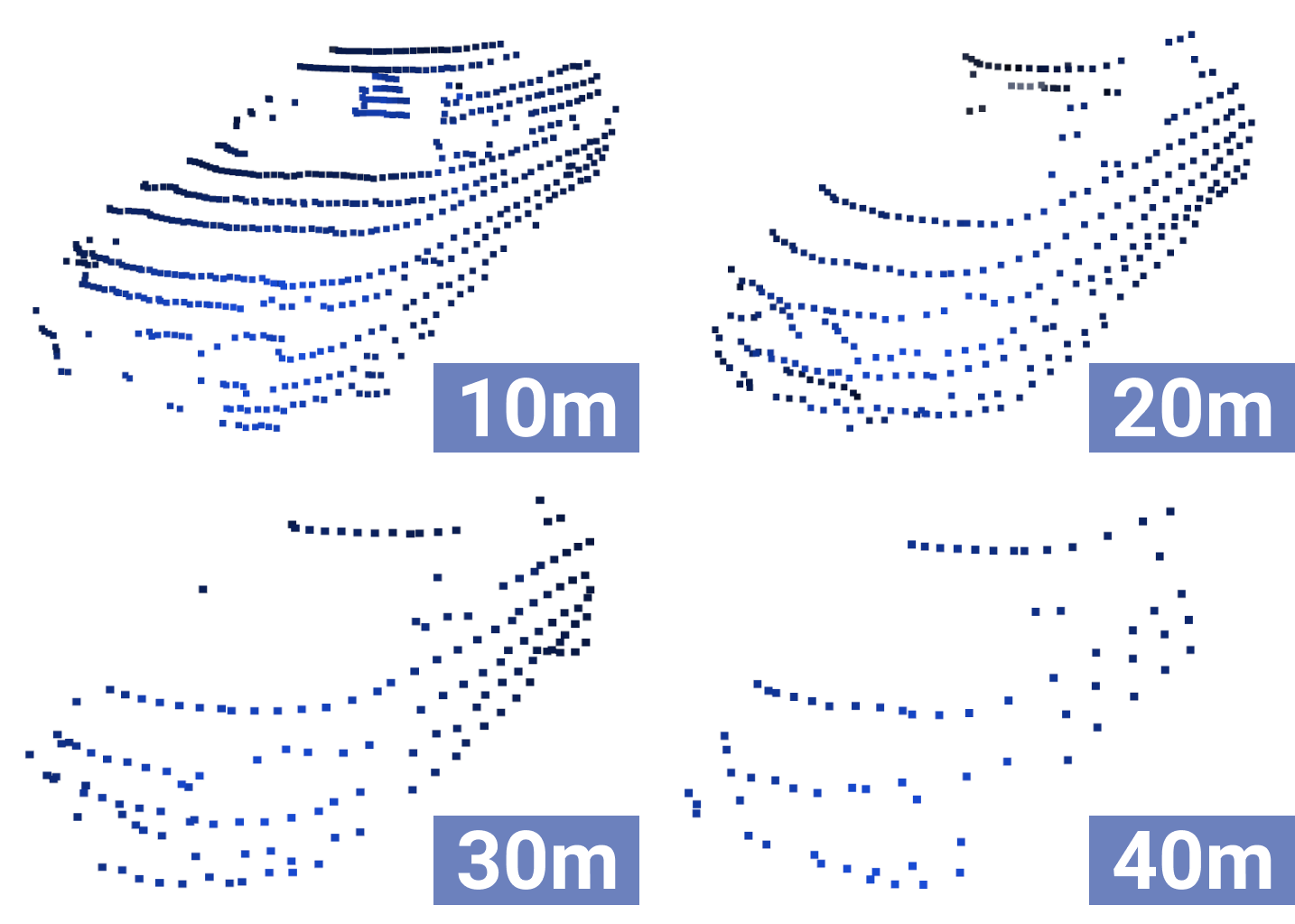}
    \caption{}
    \label{fig:ring_separation-b}
  \end{subfigure}
  \caption{(a) Lidar ring separation with increasing distance. At 40m, the distance between the vertical rings of the nuScenes lidar is nearly 1m; (b) Velodyne HDL-64E (KITTI) ring separation at 10m, 20m, 30m and 40m. }
  \label{fig:ring_separation}
  \vspace{-4mm}
\end{figure}

\noindent
{\bf VRES-based Clustering.} To find the points of the $k$-th object $o_k^T$ from the point set $s_k^T$ of the $i$-th point cloud, we employ a DB-Scan \cite{ester1996density} clustering strategy using an estimated vertical resolution (VRES) of the lidar. A typical characteristic of lidars is that the separation between points widens with distance. For ring-based lidars, such as KITTI, this widening is more prevalent in the vertical separation between points of different rings as illustrated in \Cref{fig:ring_separation}. Therefore, a good clustering distance $\epsilon$ needs to be adaptive to the widening of vertical point distances. We can estimate a VRES $\phi_{v}$ with $\phi_{v} = \frac{\theta_F}{N_R}$, where $\theta_F$ is the vertical field-of-view (FOV) of the lidar and $N_R$ is the number of rings. For a non-ring based lidar, $N_R$ was approximated based on the observed resolution. The clustering distance $\epsilon$ can be set by  $\epsilon = \alpha d_v$, where $\alpha$ is a scaling factor for the vertical point distance $d_v = d_o\text{tan}(\phi_v)$ and $d_o = ||p^T_k||_2$ is the centroid distance of the mask points. We select the largest cluster of $p^T_k$ as the set of our object points $o^T_k$. 

\subsection{Surface Completion}
\label{sec:phase_surface}
To recover the geometry of the isolated object points $o^S_k$ and $o^T_k$, we use the Ball-Pivoting Algorithm (BPA) \cite{bernardini1999ball} to interpolate a triangle mesh. The core idea of BPA is simple. A set of spheres are rolled around the point set. If any ball touches three points and doesn't fall through, it forms a triangle. We estimate surface normals for the object point set by orienting them towards the lidar origin. BPA uses these surface normals to ensure that triangles are only generated on the surface of the object; this was effective in recovering an accurate geometry of the object. The added benefit of using an SC method is that we address the issue of partial occlusion, where objects are split into two components, most commonly in driving datasets by poles, signs or trees. With BPA we can connect the two disjointed components into a single object. BPA can be easily swapped out with other SC methods such as Alpha Shapes \cite{edelsbrunner1983shape} or Poisson Surface Reconstruction \cite{kazhdan2006poisson}, which do not require pose normalization \cite{yuan2018pcn, yu2021pointr}. We explore this further in \cref{sec:ablation_studies}.

\subsection{Point Sampling}
\label{sec:phase_sampling}
Objects at closer ranges typically have more confident detections due to the high density of points; this can be observed in the Waymo leaderboard \cite{sun2020scalability} when scoring by range. In the last phase of SEE, we choose to upsample every object's triangle mesh to increase the confidence of 3D detectors. We sample the triangle meshes using Poisson disk sampling \cite{yuksel2015sample}, where each point is approximately equidistant to neighbouring points. 

To increase the confidence of detections at a further range, we upsample the meshes to emulate the point density of objects at a closer range (5-10m). Smaller point distances are typically correlated with a higher density of points. We define an ideal vertical point distance $d_\text{ideal}$, as our desired density of points. The relationship between $d_\text{ideal}$ and vertical point distance $d_v$ can be represented as $d_\text{ideal} = \frac{d_v}{\beta}$, since $d_v$ increases with distance. By rearranging this equation, we can solve for our upsampling factor $\beta$. Whilst more points are generally better, densely sampled points may accentuate mistakes in the object isolation phase. We sample each $i$-th point cloud object according to this strategy to obtain our semi-canonical representation $\mathbf{M}^S_i$ and $\mathbf{M}^T_i$.

\subsection{Object Detection in the Semi-Canonical Domain}
We train 3D object detectors on the semi-canonical domain $\rho(\mathbf{M}^S_i)$ using labels in the source domain $\mathbf{Y}^S_i$. We demonstrate the framework on two SoTA detectors SECOND-IoU \cite{yang2021st3d} and PV-RCNN, which required no architectural modifications. Given a point cloud in the test domain $\mathbf{X}^T_i$ we isolate the objects $\mathbf{O}^T_i$; convert to the semi-canonical domain $\mathbf{M}^T_i$; apply the detection network $\rho(\mathbf{M}^T_i)$; and finally, project detections back into the target domain $\mathbf{X}^T_i$.

\section{Experiments}
In this section, we first validate SEE on public datasets and subsequently demonstrate its efficacy on the novel Baraja Spectrum-Scan™ dataset. 

\subsection{Datasets}
\label{sec:experiment_datasets}
We evaluate our framework on the ``Car" class using three public datasets: Waymo, KITTI and nuScenes; and a manually annotated Baraja Spectrum-Scan™ dataset. We demonstrate the concept across multiple scenarios: (1) Differing lidar ring numbers (i.e., nuScenes $ \rightarrow $ KITTI); (2) Multiple concatenated point clouds to a single point cloud (i.e., Waymo $ \rightarrow $ KITTI); (3) Ring-based to a uniform, interleaved scan pattern (i.e., Waymo/nuScenes $ \rightarrow $ Baraja). We evaluate all scenarios on SECOND-IoU and PV-RCNN. 

\noindent
{\bf KITTI.} The KITTI dataset \cite{geiger2012we} provides 7,481 training samples and 7,518 testing samples. The entire training dataset contains 27,459 cars. The sensor suite includes 2 colour cameras and a Velodyne HDL64E lidar with 360\textdegree \space horizontal FOV (HFOV) and a narrow 26.8\textdegree \space vertical FOV (VFOV) with 64 equally spaced angular subdivisions (approximately 0.4\textdegree). We follow the popular 50/50 training/validation split \cite{chen2017multi, yan2018second}, giving 3,712 training samples and 3,769 validation samples. 

\noindent
{\bf nuScenes.} The nuScenes dataset \cite{caesar2020nuscenes} provides 28,130 training, 6,019 validation and 6,008 testing samples. The nuScenes sensor suite includes 6 cameras and a Velodyne HDL32E lidar with 360\textdegree \space HFOV and a wide 40\textdegree \space VFOV with 32 equally spaced angular subdivisions (approximately 1.25\textdegree). For fair comparison across datasets, and due to limited hardware resources, we created a subset of nuScenes comprising of 4025 training samples, and 3980 validation samples. This subset contains a similar number of cars and frames as in KITTI. To overcome the surface completion challenges caused by point cloud sparsity, we truncated the point cloud range in the training process to [-24, 24]m for X and Y, and [-2,4]m for Z axis.

\noindent
{\bf Waymo.} The Waymo Open Dataset \cite{sun2020scalability} contains 1,000 scenes, each with 200 frames. The dataset provides a top-mounted long-range lidar and 4 short-range lidars, with the point cloud truncated to a maximum of 75m for the top-lidar and 20m for the short-range lidars. The 360\textdegree \space HFOV top-lidar has a narrow VFOV of 20\textdegree \space with 64 non-uniformly spaced subdivisions where points are denser at the horizon. We concatenate all 5 lidar scans into a single point cloud. Similar to nuScenes, we create a subset of the Waymo dataset. Selecting one frame from each scene leads to a subset of 1,000 training samples with a similar number of cars (30,405 cars) to the KITTI dataset. We trained the models on the ``Vehicle" class with a point cloud range of [-75.2, 75.2]m for X and Y axis, and [-2,4]m for Z axis. 

\begin{figure}[t]
  \centering
   \includegraphics[width=0.99\linewidth]{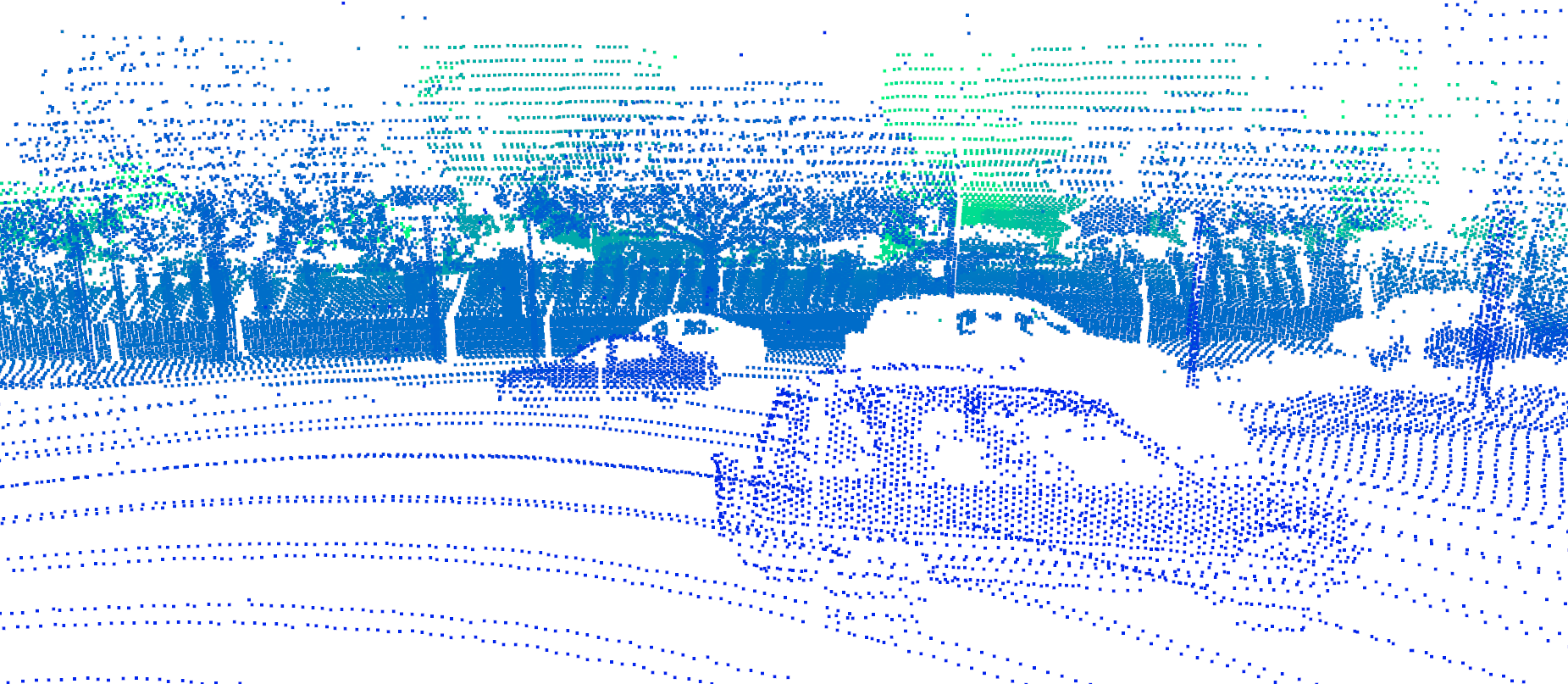}
   \caption{The Baraja Spectrum-Scan™ dataset scan pattern is configured to foveate points at the horizon. The foveation pattern is visible on the SUV where the bottom half has a ring-like appearance whilst the upper half we describe as a uniform, interleaved scan pattern.}
   \label{fig:baraja_dataset}
   \vspace{-4mm}
\end{figure}

\noindent
{\bf Baraja Spectrum-Scan™.} Our manually labelled dataset uses Baraja Spectrum-Scan™, a novel, high-resolution lidar that can dynamically increase point cloud resolution around key objects \cite{baraja2020next}. Baraja Spectrum-Scan™ has 120\textdegree \space HFOV and 30 \textdegree \space VFOV. It is able to adjust its vertical (from 0.025\textdegree-0.2\textdegree) and horizontal (from 0.05\textdegree-0.2\textdegree) angular resolution for user-defined, non-uniformly spaced subdivisions. This allows it to have a scan pattern that can achieve up to 1000 subdivisions per beam channel. Our collected dataset uses 128 non-uniform subdivisions per channel, with the points foveated to focus on the horizon as shown in \Cref{fig:baraja_dataset}. In our particular configuration, the foveated region is what we describe as a uniform, interleaved distribution of points, where the vertical and horizontal angular resolution is approximately equal. The top and bottom of the vertical FOV, however, is configured similarly to the beams of KITTI, Waymo, and nuScenes. Our camera, Intel RealSense D435i, was mounted directly below the lidar. Using the labelling tool, SUSTechPOINTS \cite{li2020sustech}, we followed the KITTI annotation scheme and labelled cars only if they are visible in the image FOV. In the image extremities, we only label cars if more than 50\% of the car is visible in the image. We adopt the same dataset limitations as Waymo in truncating the point cloud to a maximum of 75m, and we include side mirrors within the annotated bounding box. The dataset was collected in San Francisco in dry weather and contains 100 frames, with a total of 928 cars.

\subsection{SEE Results on Public Datasets}
\label{sec:experiment_public_results}
\noindent
{\bf Comparisons.} SEE was compared in the UDA scenario where no prior information about the target domain is known. We adopt a similar convention to \cite{yang2021st3d} and compare SEE with (1) Source-only, where no DA strategies are used; (2) ST3D \cite{yang2021st3d}, the SoTA UDA method on 3D object detection using self-training. Trained ST3D models were obtained from their open-sourced codebase; (3) SEE-Ideal, where we use the ground truth annotations to isolate the target domain objects; (4) Oracle, the fully supervised detector trained on the target domain. For (3), the ideal case highlights the potential of this framework with further improvements to object isolation, viewpoint alignment and image instance segmentation in the target domain.

\noindent
{\bf Evaluation Metric.} We evaluate SEE on the ``Car" class in the KITTI validation dataset, similar to other UDA methods \cite{yang2021st3d, wang2020train}. We follow the official KITTI evaluation metric and report the average precision (AP) over 40 recall positions at the 0.7 IoU threshold for 3D and BEV. We report the performance increase of models trained with SEE, over the Source-only. We evaluate all methods only on ground truth annotations where there are 50 points or more within the 3D bounding box. This is because when there are less than 50 points, it is often insufficient to generate a significant triangle mesh \cite{stutz2018learning}. We explore this further in \Cref{sec:ablation_studies}. 

\noindent
{\bf Implementation Details.} We use the Hybrid Task Cascade (HTC) instance segmentation network \cite{chen2019hybrid} provided by the mmdetection codebase \cite{mmdetection} which was pretrained on the COCO dataset \cite{lin2014microsoft}. In our SEE framework, we set $d_\text{ideal}=0.05$, $\alpha=5$, 20 BPA spheres with upper radius $1.155$, and we shrink instance segmentation masks by 2\%. We use the Open3D\cite{zhou2018open3d} implementation of BPA. For cross-dataset training and evaluation, we adopted the settings of the ST3D \cite{yang2021st3d} codebase which was based on OpenPCDet \cite{openpcdet2020}. We add an offset to the z coordinates of nuScenes (+1.8m), KITTI (+1.6m), and Baraja Spectrum-Scan™ (+2.2m) to shift the origin to the ground plane. For training, we use the widely adopted data augmentation strategies: random flipping, random world scaling, random world rotation, and random object scaling (proposed by \cite{yang2021st3d}). All models were trained on an RTX 2080 Ti.

\begin{table*}
\centering
\begin{tabular}{c|c|c|c|c|c} 
\toprule
\multirow{2}{*}{Task}             & \multirow{2}{*}{Method} & \multicolumn{2}{c|}{SECOND-IoU}                   & \multicolumn{2}{c}{PV-RCNN}                        \\ 
\cline{3-6}
                                  &                         & $\text{AP}_\text{BEV}$ / $\text{AP}_\text{3D}$                       & Improvement     & $\text{AP}_\text{BEV}$ / $\text{AP}_\text{3D}$                        & Improvement      \\ 
\hline
\multirow{5}{*}{Waymo $ \rightarrow $ KITTI}    & Source-only             & 47.67 / 11.92                   & -               & 67.42 / 41.54                   & -                \\
                                  & ST3D                    & 78.93 / 62.81                   & +31.26 / +50.88 & 89.94 / 73.32                   & +22.53 / +31.78  \\ 
\cline{2-6}
                                  & SEE                     & \textbf{88.09} /\textbf{ 65.52} & +40.42 / +53.60 & \textbf{90.82 }/\textbf{ 79.39} & +23.41 / +37.85  \\
                                  & Ideal-SEE               & 88.93 / 70.48                   & +41.26 / +58.56 & 93.17 / 84.49                   & +25.75 / +42.95  \\ 
\cline{2-6}
                                  & Oracle                  & 86.07 / 81.73                   & -               & 92.17 / 89.32                   & -                \\ 
\hline\hline
\multirow{5}{*}{Waymo $ \rightarrow $ nuScenes} & Source-only             & 21.38 / 07.29                   & -               & 29.59 / 15.09                   & -                \\
                                  & ST3D                    & \textbf{38.48} / \textbf{21.48} & +17.10 / +14.19 & \textbf{38.70} / \textbf{19.10} & +09.11 / +04.01  \\ 
\cline{2-6}
                                  & SEE                     & 21.28 / 10.51                   & -00.11 / +03.22 & 23.84 / 14.72                   & -05.75 / -00.36  \\
                                  & Ideal-SEE               & 24.27 / 12.20                   & +02.88 / +04.91 & 27.36 / 18.41                   & -02.23 / +03.32  \\ 
\cline{2-6}
                                  & Oracle                  & 26.15 / 22.37                   & -               & 26.39 / 24.83                   & -                \\ 
\hline\hline
\multirow{5}{*}{nuScenes $ \rightarrow $ KITTI} & Source-only             & 48.74 / 16.39                   & -               & 68.97 / 48.03                   & -                \\
                                  & ST3D                    & 76.82 / \textbf{58.58}          & +28.08 / +42.19 & 78.36* / 70.85*                   & +09.39 / +22.82  \\ 
\cline{2-6}
                                  & SEE                     & \textbf{77.00} / 56.00          & +28.26 / +39.61 & \textbf{85.53} / \textbf{72.51} & +16.56 / +24.49  \\
                                  & Ideal-SEE               & 83.80 / 66.82                   & +35.06 / +50.43 & 89.94 / 80.29                   & +20.98 / +32.26  \\ 
\cline{2-6}
                                  & Oracle                  & 86.07 / 81.73                   & -               & 92.17 / 89.32                   & -                \\
\bottomrule
\end{tabular}
\caption{Results for the different DA scenarios. We report the AP for the car class on KITTI dataset in bird's-eye-view ($\text{AP}_{\text{BEV}}$) and 3D ($\text{AP}_{\text{3D}}$) at the 0.7 IoU threshold on the moderate case. ``Improvement" denotes the AP increase from source-only to the various approaches in the format $\text{AP}_\text{BEV}$ / $\text{AP}_\text{3D}$. The best results are indicated in bold (with the exception of SEE-Ideal as it uses ground truth boxes). All models are evaluated on ground truth boxes with 50 points or more. Publicly available ST3D models are evaluated with this same criterion. *nuScenes → KITTI results for ST3D with PV-RCNN are from the original paper, evaluated on all ground-truth boxes as the model was not available from the codebase at the time.}
\label{tab:main_results}
\vspace{-2mm}
\end{table*}

\begin{table}
\centering
\begin{tabular}{c|c|ll|ll} 
\toprule
\multirow{3}{*}{Source}   & \multirow{3}{*}{Method} & \multicolumn{2}{c|}{SECOND-IoU}                          & \multicolumn{2}{c}{PV-RCNN}                              \\  
\cline{3-6}
                          &                         & \multicolumn{1}{c}{3D}     & \multicolumn{1}{c|}{BEV}    & \multicolumn{1}{c}{3D}     & \multicolumn{1}{c}{BEV}     \\ 
\hline
\multirow{3}{*}{nuScenes} & Source-only             & 1.02                       & 5.12                        & 10.85                      & 13.74                       \\
                          & SEE                     & 34.54                      & 58.45                       & 64.34                      & 77.73                       \\
\cline{2-6}                          
                          & Improvement             & \multicolumn{1}{c}{\textcolor[rgb]{0,0.502,0}{\textbf{+33.52}}} & \multicolumn{1}{c|}{\textcolor[rgb]{0,0.502,0}{\textbf{+53.33}}} & \multicolumn{1}{c}{\textcolor[rgb]{0,0.502,0}{\textbf{+53.49}}} & \multicolumn{1}{c}{\textcolor[rgb]{0,0.502,0}{\textbf{+63.99}}}  \\ 
\hline\hline
\multirow{3}{*}{Waymo}    & Source-only             & 49.96                      & 74.64                       & 76.14                      & 84.10                       \\
                          & SEE                     & 73.79                      & 84.74                       & 79.13                      & 87.79                       \\
\cline{2-6}                          
                          & Improvement             & \multicolumn{1}{c}{\textcolor[rgb]{0,0.502,0}{\textbf{+23.84}}} & \multicolumn{1}{c|}{\textcolor[rgb]{0,0.502,0}{\textbf{+10.10}}} & \multicolumn{1}{c}{\textcolor[rgb]{0,0.502,0}{\textbf{+2.98}}}  & \multicolumn{1}{c}{\textcolor[rgb]{0,0.502,0}{\textbf{+3.69}}}   \\
\bottomrule
\end{tabular}
\caption{Results for the Baraja Spectrum-Scan™ dataset. ``Improvement" denotes AP increase from Source-only to SEE. Reported AP is at the 0.7 IoU threshold. We evaluate using the same trained models in \Cref{tab:main_results}.}
\label{tab:baraja_results}
\vspace{-4mm}
\end{table}

\noindent
{\bf Results.} 
From \Cref{tab:main_results}, we show that SEE improves the performance of both detectors by a large margin on all unsupervised DA scenarios. Due to the nature of our approach, having more points leads to a better-interpolated geometry. This is observed in the Waymo $ \rightarrow $ KITTI task, where our approach outperforms ST3D with both SECOND-IoU and PV-RCNN by 2.71 and 6.07 $\text{AP}_\text{3D}$ respectively. From this, we demonstrate that SEE, an MTDA method, is able to achieve similar or higher performance than the SoTA single target DA method, ST3D. For the 64-beam KITTI/Waymo lidars, 20-50 points on an object often outlines 3D geometry as the points are spread across more than two rings. In nuScenes however, due to having less beams and large vertical angular resolution, 20-50 points are often in the form of 1-2 rings which reduces the point set's dimension to a line, or a plane. Whilst each ring may be dense horizontally and have around 50 points for 2 rings on a car, a line or plane does not give any 3D information. The lack of 3D geometry in the object points poses a challenging task when nuScenes is the source or target domain, causing failed or poor interpolation of object geometry. The result of this is minor performance gain or even degradation in the Waymo $ \rightarrow $ nuScenes task. This is also apparent in nuScenes $ \rightarrow $ KITTI, where SEE achieves comparable performance to ST3D with SECOND-IoU. We additionally note that there is a substantial gap when comparing SEE to SEE-Ideal, which indicates the potential of our chosen SC method, BPA, with improvements to the object isolation phase and future instance segmentation models. We believe that the gap between SEE and SEE-Ideal can be further closed if the camera and lidar viewpoint alignment is minimised, as this reduces background points. Whilst we do not explicitly deal with the bounding box distribution gap \cite{wang2020train}, it is an inherent benefit of SEE as the surface completed objects provides the detector with better prior estimates, leading to better car size predictions.

\subsection{SEE for Baraja Spectrum-Scan™}
\label{sec:experiment_baraja_results}
In \Cref{tab:baraja_results}, we highlight SEE's adoption of novel lidar technology by evaluating on the Baraja Spectrum-Scan™ dataset. We demonstrate that the same trained models in \Cref{tab:main_results} can attain high performance in an unseen dataset with a distinct scan pattern (see \Cref{fig:baraja_dataset}). When comparing with nuScenes, due to its ring-like nature and sparsity of beams, the scan pattern of the lidar is vastly different. Even at a close range, a car in nuScenes bears little resemblance to any car perceived by Baraja Spectrum-Scan™. In \Cref{tab:baraja_results}, this is observed in the dramatically low performance of models trained on nuScenes as the source domain, obtaining 1.02 and 10.85 in $\text{AP}_\text{3D}$ for SECOND-IoU and PV-RCNN respectively. When the same models are trained with SEE, the performance is drastically improved with an increase of 1.02 $ \rightarrow $ 34.54 and 10.85 $ \rightarrow $ 64.34 in $\text{AP}_\text{3D}$. For Waymo, lidar beams are clustered in the middle of vertical angles, leading to high resolution at the horizon and sparse rings in the top and bottom of the FOV, similar to Baraja Spectrum-Scan™. This scan pattern similarity leads to a higher performing ``Source-only". We observe that for the Waymo ``Source-only" PV-RCNN attains a high $\text{AP}_\text{3D}$ for the Baraja dataset, however the same ``Source-only" model does not perform well on KITTI as shown in \Cref{tab:main_results}. The single SEE-trained PV-RCNN model however, can obtain high performance on both KITTI and Baraja Spectrum-Scan™. In \Cref{tab:baraja_results} we show that SEE further pushes SECOND-IoU and PV-RCNN's performance to 73.79 and 79.13 $\text{AP}_\text{3D}$ respectively. 

\section{Ablation Studies}
\label{sec:ablation_studies}
In this section, we conduct ablation studies using the SECOND-IoU model. 

\begin{figure}[t]
  \centering
  \begin{subfigure}{0.49\linewidth}
    \includegraphics[width=0.99\linewidth]{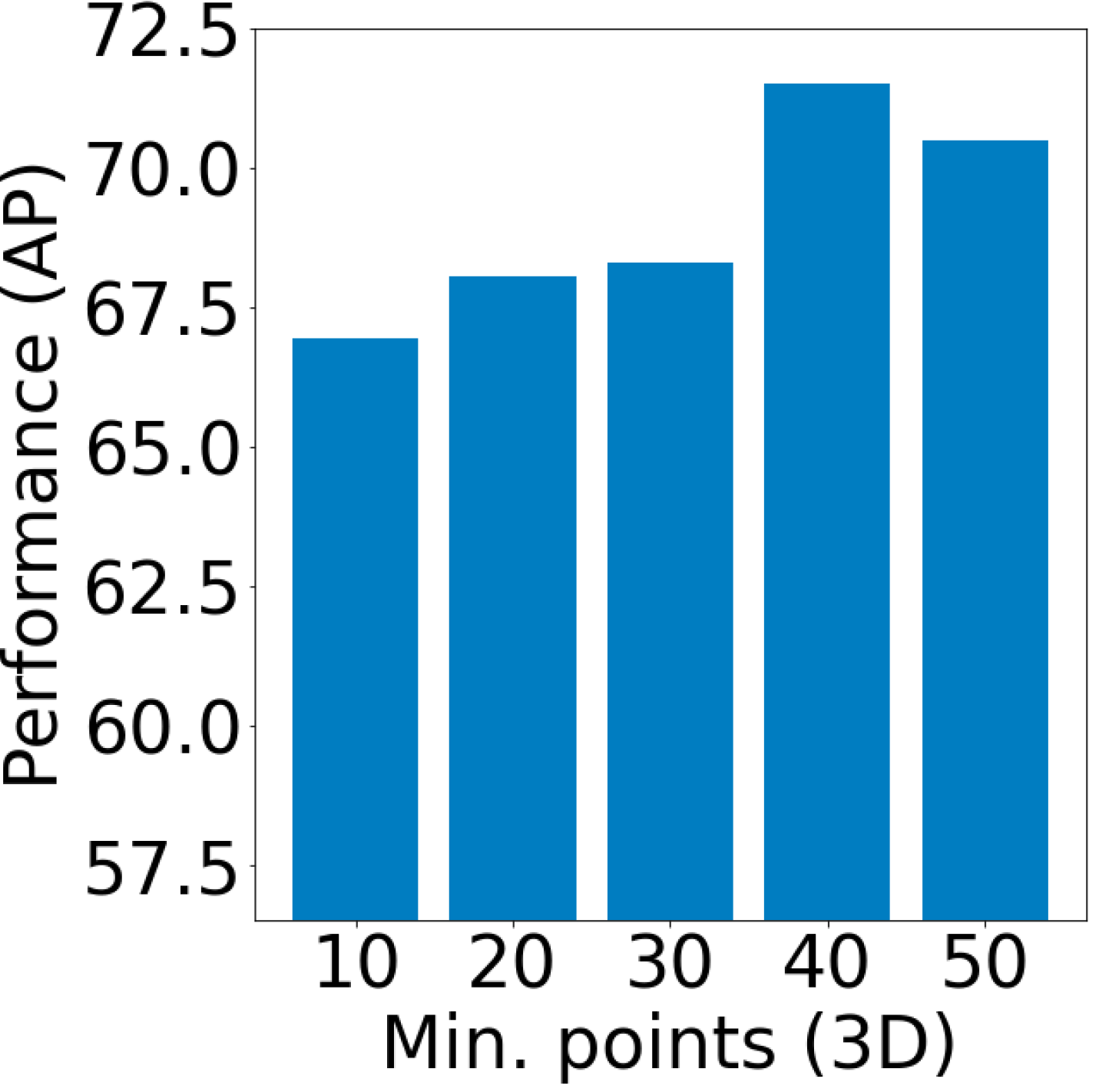}
    \caption{}
    \label{fig:min_pts_3d}
  \end{subfigure}
  \hfill
  \begin{subfigure}{0.49\linewidth}
    \includegraphics[width=0.99\linewidth]{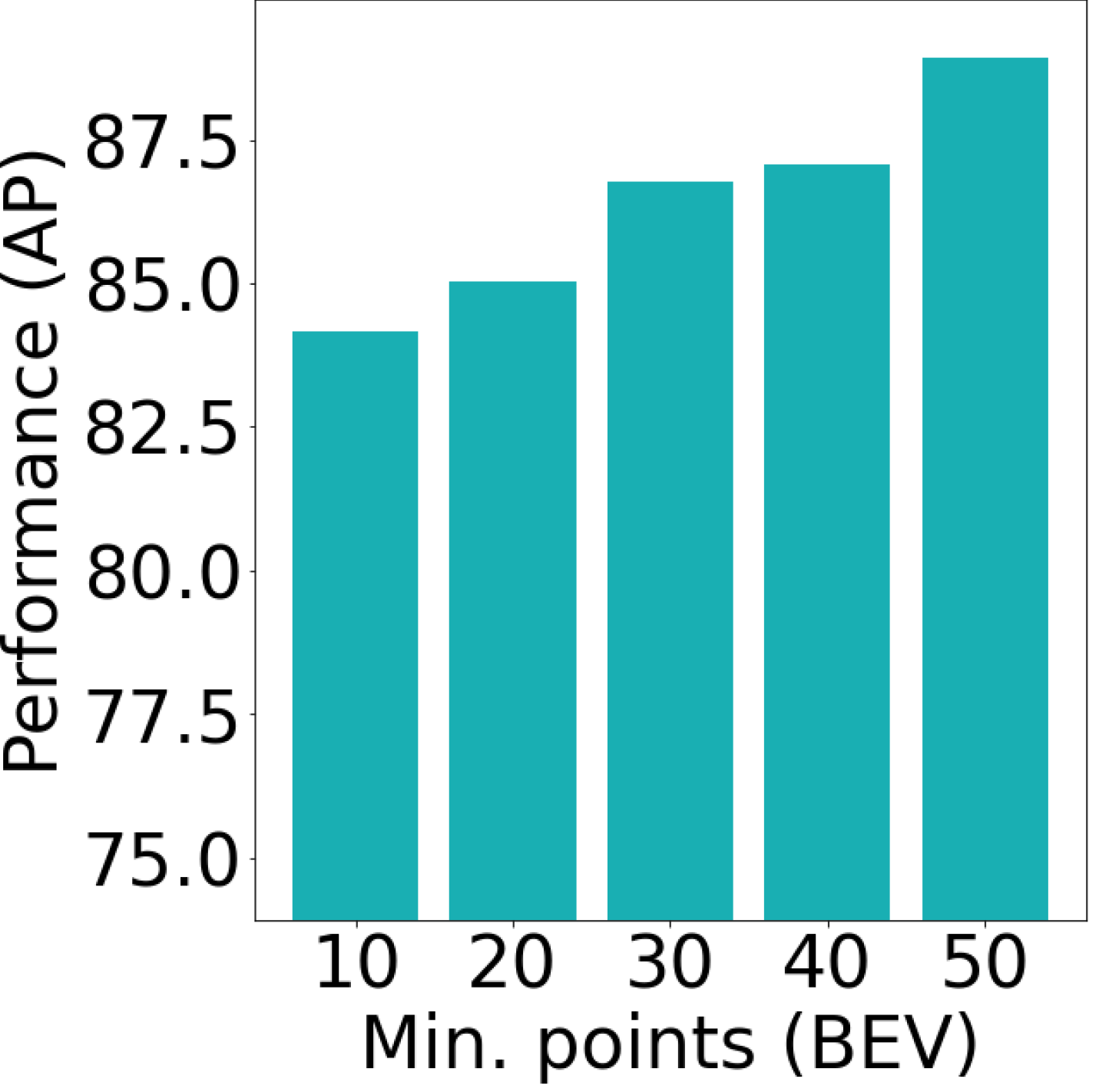}
    \caption{}
    \label{fig:min_pts_bev}
  \end{subfigure}
  \caption{Minimum number of points using SEE-Ideal, evaluated on the Waymo $\rightarrow $ KITTI scenario with SECOND-IoU. Reported AP is at 0.7 IoU for (a) 3D (b) BEV.}
  \label{fig:min_pts}
\end{figure}

\noindent
{\bf Minimum number of points.} To interpolate the underlying geometry, we need to have a sufficient amount of points to interpolate from. We investigate the performance of SEE on objects containing less than 50 points and adjust the evaluation accordingly. For example, if the minimum number of points for the SC phase is 40, we also evaluate only on ground-truth boxes that have 40 points or more. It can be seen in \Cref{fig:min_pts} that more points leads to higher quality meshes, and hence, better detection performance.

\begin{table}
\centering
\begin{tabular}{c|c|cc} 
\toprule
Strategy                                     & Config & 3D    & BEV    \\ 
\hline
\multirow{3}{*}{VRES ($d_\text{ideal}$)}   & 0.07   & 64.13 / 58.70 & 85.70 / 82.36  \\
                                           & 0.05   & 70.96 / 65.52 & 89.70 / 88.09  \\
                                           & 0.03   & 68.62 / 57.16 & 89.50 / 87.94  \\ 
\hline
\multirow{3}{*}{SA ($\frac{\text{pts}}{m^2}$)} & 300    & 70.86 / 58.70 & 89.67 / 80.68  \\
                                           & 500    & 75.52 / 61.41 & 91.17 / 80.34  \\
                                           & 700    & 73.86 / 61.80 & 89.63 / 81.09  \\ 
\hline
VL                             & -      & 70.02 / 64.23 & 86.71 / 85.57  \\
\bottomrule
\end{tabular}
\caption{Point sampling strategies: VRES, Surface-Area (SA), and Virtual Lidar (VL). The reported AP shows SEE-Ideal/SEE at 0.7 IoU using SECOND-IoU on the Waymo $ \rightarrow $ KITTI task. Methods are evaluated on ground truth boxes with 50 or more points.}
\label{tab:sampling}
\vspace{-4mm}  
\end{table}

\noindent
{\bf Point Sampling Strategy.} We compare our VRES-based sampling with two other strategies: (1) Surface-area (SA) based sampling, where we sample a certain number of points per $m^2$ of the triangle mesh surface area; (2) Virtual lidar (VL), where we adopt the strategy of \cite{yi2021complete} mimicking the scan pattern of the target domain by comparing the spherical coordinates of the point clouds. In \Cref{tab:sampling} we can see that a increasing the density of point sampling in $d_\text{ideal}=0.03$ and $\frac{\text{pts}}{m^2}=700$ causes a decrease in performance, as it potentially over-accentuates mistakes of the SC method. On the flip side, fewer points lead to a weaker presence of the object, which also decreases the performance. 

\noindent
{\bf Surface completion.} In \Cref{tab:mesh_methods}, we compare another SC method, alpha shapes \cite{edelsbrunner1983shape} and a deep learning (DL) shape completion method, PCN \cite{yuan2018pcn}. We notice that alpha shapes tend to join the interior object points, leading to irregularly shaped cars, which is shown in the performance degradation in $\text{AP}_\text{3D}$ when comparing with BPA. With PCN, we demonstrate the potential of our framework when using DL shape completion methods. In this table, we only use PCN with SEE-Ideal as we require the ground truth boxes to normalise the pose and scale of each car for completion. 

\section{Discussion and Future Work}

Whilst high-performing, our framework does not run in real-time. The SC phase of SEE runs at 0.33FPS for a single camera-lidar pair on an Intel(R) Core(TM) i7-6700K CPU. As the primary intent for SEE is for offline unified detector evaluation, these processing times are acceptable. We showed in \Cref{fig:min_pts} that our method performs better with more points on the object. This is also reflected in the high performance of SEE across lidars with at least 64 beams, and lower performance when using 32-beam lidars. However, we consider this a minor limitation as we argue that the future direction of lidar manufacturing is headed towards higher resolution point clouds. The results in \Cref{tab:mesh_methods} indicates the importance of the SC phase and leads us to believe that better SC methods can further improve the domain adaptation capabilities of SEE. We believe that there is potential in our framework to adopt DL shape completion methods \cite{yu2021pointr, yuan2018pcn} in our SC phase. The difficulty in naively adopting these methods, however, is that they require the ground truth boxes to normalise the object pose and scale. Incorrect normalization can cause the shape completion network to generate spurious cars in odd rotations and scales which is detrimental to the detector. Future work will include testing SEE on other classes and researching shape completion methods without the use of scale and pose supervision. 

\begin{table}
\centering
\begin{tabular}{l|cc} 
\toprule
\multicolumn{1}{c|}{Method} & 3D             & BEV             \\ 
\hline
Ball Pivoting               & 66.81/55.78    & 82.67/76.80     \\ 
Alpha Shapes                & 56.31/49.07    & 68.57/65.17     \\ 
PCN \cite{yuan2018pcn}      & 83.89/~ ~-~ ~ & 85.60/~ ~-~ ~  \\
\bottomrule
\end{tabular}
\caption{Comparison of BPA with Alpha Shapes and PCN \cite{yuan2018pcn}. Table shows SEE-Ideal/SEE on the nuScenes $ \rightarrow $ KITTI task using SECOND-IoU for $\text{AP}_{\text{3D}}$ at 0.7 IoU. }
\label{tab:mesh_methods}
\vspace{-4mm}  
\end{table}

\section{Conclusion}

We present SEE, a lidar-agnostic 3D object detection framework to enable any 3d detector to be lidar-agnostic without manual labelling or fine-tuning of the model. We demonstrate that the scan pattern discrepancy is a large component of the domain gap. Extensive experiments on public datasets and a novel lidar dataset validate the efficacy of our MTDA framework. Furthermore, through this, we show that our MTDA framework can attain competitive performance against the single-target domain adaptation state of the art method. 


\bibliography{IEEEabrv, references}

\begin{thebibliography}{10}
\providecommand{\url}[1]{#1}
\csname url@rmstyle\endcsname
\providecommand{\newblock}{\relax}
\providecommand{\bibinfo}[2]{#2}
\providecommand\BIBentrySTDinterwordspacing{\spaceskip=0pt\relax}
\providecommand\BIBentryALTinterwordstretchfactor{4}
\providecommand\BIBentryALTinterwordspacing{\spaceskip=\fontdimen2\font plus
\BIBentryALTinterwordstretchfactor\fontdimen3\font minus
  \fontdimen4\font\relax}
\providecommand\BIBforeignlanguage[2]{{%
\expandafter\ifx\csname l@#1\endcsname\relax
\typeout{** WARNING: IEEEtran.bst: No hyphenation pattern has been}%
\typeout{** loaded for the language `#1'. Using the pattern for}%
\typeout{** the default language instead.}%
\else
\language=\csname l@#1\endcsname
\fi
#2}}

\bibitem{sun2020scalability}
P.~Sun, H.~Kretzschmar, X.~Dotiwalla, A.~Chouard, V.~Patnaik, P.~Tsui, J.~Guo,
  Y.~Zhou, Y.~Chai, B.~Caine, \emph{et~al.}, ``Scalability in perception for
  autonomous driving: Waymo open dataset,'' in \emph{Proceedings of the
  IEEE/CVF Conference on Computer Vision and Pattern Recognition}, 2020, pp.
  2446--2454.

\bibitem{geiger2012we}
A.~Geiger, P.~Lenz, and R.~Urtasun, ``Are we ready for autonomous driving? the
  kitti vision benchmark suite,'' in \emph{2012 IEEE conference on computer
  vision and pattern recognition}.\hskip 1em plus 0.5em minus 0.4em\relax IEEE,
  2012, pp. 3354--3361.

\bibitem{caesar2020nuscenes}
H.~Caesar, V.~Bankiti, A.~H. Lang, S.~Vora, V.~E. Liong, Q.~Xu, A.~Krishnan,
  Y.~Pan, G.~Baldan, and O.~Beijbom, ``nuscenes: A multimodal dataset for
  autonomous driving,'' in \emph{Proceedings of the IEEE/CVF conference on
  computer vision and pattern recognition}, 2020, pp. 11\,621--11\,631.

\bibitem{shi2020pv}
S.~Shi, C.~Guo, L.~Jiang, Z.~Wang, J.~Shi, X.~Wang, and H.~Li, ``Pv-rcnn:
  Point-voxel feature set abstraction for 3d object detection,'' in
  \emph{Proceedings of the IEEE/CVF Conference on Computer Vision and Pattern
  Recognition}, 2020, pp. 10\,529--10\,538.

\bibitem{yang2021st3d}
J.~Yang, S.~Shi, Z.~Wang, H.~Li, and X.~Qi, ``St3d: Self-training for
  unsupervised domain adaptation on 3d object detection,'' in \emph{Proceedings
  of the IEEE/CVF Conference on Computer Vision and Pattern Recognition}, 2021,
  pp. 10\,368--10\,378.

\bibitem{yi2021complete}
L.~Yi, B.~Gong, and T.~Funkhouser, ``Complete \& label: A domain adaptation
  approach to semantic segmentation of lidar point clouds,'' in
  \emph{Proceedings of the IEEE/CVF Conference on Computer Vision and Pattern
  Recognition}, 2021, pp. 15\,363--15\,373.

\bibitem{yan2018second}
Y.~Yan, Y.~Mao, and B.~Li, ``Second: Sparsely embedded convolutional
  detection,'' \emph{Sensors}, vol.~18, no.~10, p. 3337, 2018.

\bibitem{yang2018pixor}
B.~Yang, W.~Luo, and R.~Urtasun, ``Pixor: Real-time 3d object detection from
  point clouds,'' in \emph{Proceedings of the IEEE conference on Computer
  Vision and Pattern Recognition}, 2018, pp. 7652--7660.

\bibitem{chen2017multi}
X.~Chen, H.~Ma, J.~Wan, B.~Li, and T.~Xia, ``Multi-view 3d object detection
  network for autonomous driving,'' in \emph{Proceedings of the IEEE conference
  on Computer Vision and Pattern Recognition}, 2017, pp. 1907--1915.

\bibitem{lang2019pointpillars}
A.~H. Lang, S.~Vora, H.~Caesar, L.~Zhou, J.~Yang, and O.~Beijbom,
  ``Pointpillars: Fast encoders for object detection from point clouds,'' in
  \emph{Proceedings of the IEEE/CVF Conference on Computer Vision and Pattern
  Recognition}, 2019, pp. 12\,697--12\,705.

\bibitem{pan20213d}
X.~Pan, Z.~Xia, S.~Song, L.~E. Li, and G.~Huang, ``3d object detection with
  pointformer,'' in \emph{Proceedings of the IEEE/CVF Conference on Computer
  Vision and Pattern Recognition}, 2021, pp. 7463--7472.

\bibitem{qi2017pointnet++}
C.~R. Qi, L.~Yi, H.~Su, and L.~J. Guibas, ``Pointnet++: Deep hierarchical
  feature learning on point sets in a metric space,'' \emph{Advances in neural
  information processing systems}, vol.~30, 2017.

\bibitem{vora2020pointpainting}
S.~Vora, A.~H. Lang, B.~Helou, and O.~Beijbom, ``Pointpainting: Sequential
  fusion for 3d object detection,'' in \emph{Proceedings of the IEEE/CVF
  conference on computer vision and pattern recognition}, 2020, pp. 4604--4612.

\bibitem{wang2019frustum}
Z.~Wang and K.~Jia, ``Frustum convnet: Sliding frustums to aggregate local
  point-wise features for amodal 3d object detection,'' in \emph{2019 IEEE/RSJ
  International Conference on Intelligent Robots and Systems (IROS)}.\hskip 1em
  plus 0.5em minus 0.4em\relax IEEE, 2019, pp. 1742--1749.

\bibitem{sankaranarayanan2018generate}
S.~Sankaranarayanan, Y.~Balaji, C.~D. Castillo, and R.~Chellappa, ``Generate to
  adapt: Aligning domains using generative adversarial networks,'' in
  \emph{Proceedings of the IEEE Conference on Computer Vision and Pattern
  Recognition}, 2018, pp. 8503--8512.

\bibitem{rozantsev2018beyond}
A.~Rozantsev, M.~Salzmann, and P.~Fua, ``Beyond sharing weights for deep domain
  adaptation,'' \emph{IEEE transactions on pattern analysis and machine
  intelligence}, vol.~41, no.~4, pp. 801--814, 2018.

\bibitem{zhang2018unsupervised}
Y.~Zhang, N.~Wang, S.~Cai, and L.~Song, ``Unsupervised domain adaptation by
  mapped correlation alignment,'' \emph{IEEE Access}, vol.~6, pp.
  44\,698--44\,706, 2018.

\bibitem{kang2019contrastive}
G.~Kang, L.~Jiang, Y.~Yang, and A.~G. Hauptmann, ``Contrastive adaptation
  network for unsupervised domain adaptation,'' in \emph{Proceedings of the
  IEEE/CVF Conference on Computer Vision and Pattern Recognition}, 2019, pp.
  4893--4902.

\bibitem{hong2018conditional}
W.~Hong, Z.~Wang, M.~Yang, and J.~Yuan, ``Conditional generative adversarial
  network for structured domain adaptation,'' in \emph{Proceedings of the IEEE
  Conference on Computer Vision and Pattern Recognition}, 2018, pp. 1335--1344.

\bibitem{wu2019squeezesegv2}
B.~Wu, X.~Zhou, S.~Zhao, X.~Yue, and K.~Keutzer, ``Squeezesegv2: Improved model
  structure and unsupervised domain adaptation for road-object segmentation
  from a lidar point cloud,'' in \emph{2019 International Conference on
  Robotics and Automation (ICRA)}.\hskip 1em plus 0.5em minus 0.4em\relax IEEE,
  2019, pp. 4376--4382.

\bibitem{jaritz2020xmuda}
M.~Jaritz, T.-H. Vu, R.~d. Charette, E.~Wirbel, and P.~P{\'e}rez, ``xmuda:
  Cross-modal unsupervised domain adaptation for 3d semantic segmentation,'' in
  \emph{Proceedings of the IEEE/CVF Conference on Computer Vision and Pattern
  Recognition}, 2020, pp. 12\,605--12\,614.

\bibitem{luo2021unsupervised}
Z.~Luo, Z.~Cai, C.~Zhou, G.~Zhang, H.~Zhao, S.~Yi, S.~Lu, H.~Li, S.~Zhang, and
  Z.~Liu, ``Unsupervised domain adaptive 3d detection with multi-level
  consistency,'' in \emph{Proceedings of the IEEE/CVF International Conference
  on Computer Vision}, 2021, pp. 8866--8875.

\bibitem{langer2020domain}
F.~Langer, A.~Milioto, A.~Haag, J.~Behley, and C.~Stachniss, ``Domain transfer
  for semantic segmentation of lidar data using deep neural networks,'' in
  \emph{2020 IEEE/RSJ International Conference on Intelligent Robots and
  Systems (IROS)}.\hskip 1em plus 0.5em minus 0.4em\relax IEEE, 2020, pp.
  8263--8270.

\bibitem{xu2021spg}
Q.~Xu, Y.~Zhou, W.~Wang, C.~R. Qi, and D.~Anguelov, ``Spg: Unsupervised domain
  adaptation for 3d object detection via semantic point generation,'' in
  \emph{Proceedings of the IEEE/CVF International Conference on Computer
  Vision}, 2021, pp. 15\,446--15\,456.

\bibitem{zhang2021srdan}
W.~Zhang, W.~Li, and D.~Xu, ``Srdan: Scale-aware and range-aware domain
  adaptation network for cross-dataset 3d object detection,'' in
  \emph{Proceedings of the IEEE/CVF Conference on Computer Vision and Pattern
  Recognition}, 2021, pp. 6769--6779.

\bibitem{qin2019pointdan}
C.~Qin, H.~You, L.~Wang, C.-C.~J. Kuo, and Y.~Fu, ``Pointdan: A multi-scale 3d
  domain adaption network for point cloud representation,'' \emph{Advances in
  Neural Information Processing Systems}, vol.~32, 2019.

\bibitem{wang2020train}
Y.~Wang, X.~Chen, Y.~You, L.~E. Li, B.~Hariharan, M.~Campbell, K.~Q.
  Weinberger, and W.-L. Chao, ``Train in germany, test in the usa: Making 3d
  object detectors generalize,'' in \emph{Proceedings of the IEEE/CVF
  Conference on Computer Vision and Pattern Recognition}, 2020, pp.
  11\,713--11\,723.

\bibitem{you2021exploiting}
Y.~You, C.~A. Diaz-Ruiz, Y.~Wang, W.-L. Chao, B.~Hariharan, M.~Campbell, and
  K.~Q. Weinberger, ``Exploiting playbacks in unsupervised domain adaptation
  for 3d object detection,'' \emph{arXiv preprint arXiv:2103.14198}, 2021.

\bibitem{Berrio2021}
J.~S. Berrio, M.~Shan, S.~Worrall, and E.~Nebot, ``Camera-lidar integration:
  Probabilistic sensor fusion for semantic mapping,'' \emph{IEEE Transactions
  on Intelligent Transportation Systems}, 2021.

\bibitem{tsai2021optimising}
D.~Tsai, S.~Worrall, M.~Shan, A.~Lohr, and E.~Nebot, ``Optimising the selection
  of samples for robust lidar camera calibration,'' in \emph{2021 IEEE
  International Intelligent Transportation Systems Conference (ITSC)}, 2021,
  pp. 2631--2638.

\bibitem{gomez2020using}
F.~Gomez-Donoso, E.~Cruz, M.~Cazorla, S.~Worrall, and E.~Nebot, ``Using a 3d
  cnn for rejecting false positives on pedestrian detection,'' in \emph{2020
  International Joint Conference on Neural Networks (IJCNN)}.\hskip 1em plus
  0.5em minus 0.4em\relax IEEE, 2020, pp. 1--6.

\bibitem{ester1996density}
M.~Ester, H.-P. Kriegel, J.~Sander, X.~Xu, \emph{et~al.}, ``A density-based
  algorithm for discovering clusters in large spatial databases with noise.''
  in \emph{kdd}, vol.~96, no.~34, 1996, pp. 226--231.

\bibitem{bernardini1999ball}
F.~Bernardini, J.~Mittleman, H.~Rushmeier, C.~Silva, and G.~Taubin, ``The
  ball-pivoting algorithm for surface reconstruction,'' \emph{IEEE transactions
  on visualization and computer graphics}, vol.~5, no.~4, pp. 349--359, 1999.

\bibitem{edelsbrunner1983shape}
H.~Edelsbrunner, D.~Kirkpatrick, and R.~Seidel, ``On the shape of a set of
  points in the plane,'' \emph{IEEE Transactions on information theory},
  vol.~29, no.~4, pp. 551--559, 1983.

\bibitem{kazhdan2006poisson}
M.~Kazhdan, M.~Bolitho, and H.~Hoppe, ``Poisson surface reconstruction,'' in
  \emph{Proceedings of the fourth Eurographics symposium on Geometry
  processing}, vol.~7, 2006.

\bibitem{yuan2018pcn}
W.~Yuan, T.~Khot, D.~Held, C.~Mertz, and M.~Hebert, ``Pcn: Point completion
  network,'' in \emph{2018 International Conference on 3D Vision (3DV)}.\hskip
  1em plus 0.5em minus 0.4em\relax IEEE, 2018, pp. 728--737.

\bibitem{yu2021pointr}
X.~Yu, Y.~Rao, Z.~Wang, Z.~Liu, J.~Lu, and J.~Zhou, ``Pointr: Diverse point
  cloud completion with geometry-aware transformers,'' in \emph{Proceedings of
  the IEEE/CVF International Conference on Computer Vision}, 2021, pp.
  12\,498--12\,507.

\bibitem{yuksel2015sample}
C.~Yuksel, ``Sample elimination for generating poisson disk sample sets,'' in
  \emph{Computer Graphics Forum}, vol.~34, no.~2.\hskip 1em plus 0.5em minus
  0.4em\relax Wiley Online Library, 2015, pp. 25--32.

\bibitem{baraja2020next}
C.~Pulikkaseril and N.~Langdale-Smith, ``Next generation lidar for a fully
  autonomous future,'' \url{https://www.baraja.com/en/technology/white-paper},
  Baraja, White Paper, 2020.

\bibitem{li2020sustech}
E.~Li, S.~Wang, C.~Li, D.~Li, X.~Wu, and Q.~Hao, ``Sustech points: A portable
  3d point cloud interactive annotation platform system,'' in \emph{2020 IEEE
  Intelligent Vehicles Symposium (IV)}, 2020, pp. 1108--1115.

\bibitem{stutz2018learning}
D.~Stutz and A.~Geiger, ``Learning 3d shape completion from laser scan data
  with weak supervision,'' in \emph{Proceedings of the IEEE Conference on
  Computer Vision and Pattern Recognition}, 2018, pp. 1955--1964.

\bibitem{chen2019hybrid}
K.~Chen, J.~Pang, J.~Wang, Y.~Xiong, X.~Li, S.~Sun, W.~Feng, Z.~Liu, J.~Shi,
  W.~Ouyang, C.~C. Loy, and D.~Lin, ``Hybrid task cascade for instance
  segmentation,'' in \emph{IEEE Conference on Computer Vision and Pattern
  Recognition}, 2019.

\bibitem{mmdetection}
K.~Chen, J.~Wang, J.~Pang, Y.~Cao, Y.~Xiong, X.~Li, S.~Sun, W.~Feng, Z.~Liu,
  J.~Xu, Z.~Zhang, D.~Cheng, C.~Zhu, T.~Cheng, Q.~Zhao, B.~Li, X.~Lu, R.~Zhu,
  Y.~Wu, J.~Dai, J.~Wang, J.~Shi, W.~Ouyang, C.~C. Loy, and D.~Lin,
  ``{MMDetection}: Open mmlab detection toolbox and benchmark,'' \emph{arXiv
  preprint arXiv:1906.07155}, 2019.

\bibitem{lin2014microsoft}
T.-Y. Lin, M.~Maire, S.~Belongie, J.~Hays, P.~Perona, D.~Ramanan,
  P.~Doll{\'a}r, and C.~L. Zitnick, ``Microsoft coco: Common objects in
  context,'' in \emph{European conference on computer vision}.\hskip 1em plus
  0.5em minus 0.4em\relax Springer, 2014, pp. 740--755.

\bibitem{zhou2018open3d}
Q.-Y. Zhou, J.~Park, and V.~Koltun, ``{Open3D}: {A} modern library for {3D}
  data processing,'' \emph{arXiv:1801.09847}, 2018.

\bibitem{openpcdet2020}
O.~D. Team, ``Openpcdet: An open-source toolbox for 3d object detection from
  point clouds,'' \url{https://github.com/open-mmlab/OpenPCDet}, 2020.

\end{thebibliography}

\addtolength{\textheight}{-12cm}   

\end{document}